# Modeling Suspicious Email Detection using Enhanced Feature Selection

Sarwat Nizamani, Nasrullah Memon, Uffe Kock Wiil, Panagiotis Karampelas

*Abstract*—The paper presents a suspicious email detection model which incorporates enhanced feature selection. In the paper we proposed the use of feature selection strategies along with classification technique for terrorists email detection. The presented model focuses on the evaluation of machine learning algorithms such as decision tree (ID3), logistic regression, Naïve Bayes (NB), and Support Vector Machine (SVM) for detecting emails containing suspicious content. In the literature, various algorithms achieved good accuracy for the desired task. However, the results achieved by those algorithms can be further improved by using appropriate feature selection mechanisms. We have identified the use of a specific feature selection scheme that improves the performance of the existing algorithms.

## I. INTRODUCTION

Email is the most popular way of communication of this era. It provides an easy and reliable method of communication. Email messages can be sent to an individual or groups. A single email can spread among millions of people within few moments. Nowadays, most individuals even cannot imagine the life without email. For those reasons, email has become a widely used medium for communication of terrorists as well. A great number of researchers [1],[2],[3],[4] focused in the area of counterterrorism after the disastrous events of 9/11 trying to predict terrorist plans from suspicious communication. This also motivated us to contribute in this area.

In this paper, we have applied data mining techniques to detect suspicious emails, i.e., an email that alerts of upcoming terrorist events. We have applied decision tree (ID3) [5], Naïve Bayes [6], logistic regression [7], and SVM [7] algorithms, emphasizing initially on feature space creation, then applying various feature selection techniques by selecting a subset of the original feature space.

Feature selection involves the choice of a feature subset

Sarwat Nizamani is at Maersk McKinney Moller Institute, University of Southern, Denmark. (email: saniz@mmmi.sdu.dk).

Nasrullah Memon is at Maersk McKinney Moller Institute, University of Southern, Denmark. (email: memon@mmmi.sdu.dk)

Uffe Kock Wiil is at Maersk McKinney Moller Institute, University of Southern, Denmark. (email: ukwiil@mmmi.sdu.dk)

Panagiotis Karampelas is at Hellenic American University, Manchester, NH, USA (email: pkarampelas@hauniv.us).

evaluator and a search method. We experimented on our dataset with various classifiers and various feature selection schemes. We also observed that for a specific classifier in choice of feature selection, an appropriate evaluator should be used with an appropriate search method. In the original feature space, we have used some keywords and some indicators. For example, if domain specific keywords are found with suspicious indicators in an email message, it is classified as suspicious, whereas the occurrence of domain specific keywords without the presence of suspicious indicators in an email, it is not classified as suspicious. With the selection of proper features, the accuracy of decision tree, SVM, Naïve Bayes and logistic regression is improved. We suggest the use of a specific feature selection scheme with a particular evaluator and an appropriate search method. In the paper, we have conducted experiments on our dataset using state of art classifiers and supervised feature selection methods. The choice of feature selection involves the selection of an evaluator and a search method. For our experiments, we applied CfsSubsetEval, ChiSquare, InfoGain, GainRatio and ConsistencySubsetEval evaluators and BestFirst, GreedyStepWise and Ranker search methods. The results show that ConsistensySubsetEval with GreedyStepWise search methods improves the performance of three out of four classifiers.

We have developed an email dataset containing suspicious emails, because there is no benchmark dataset available in the domain. Some emails in the dataset are taken from some open emails released by press concerning Mumbai attack [9]. Few of the emails are real emails of 9/11 incident which are also used by authors [1]. We also added some dummy emails resembling to terrorist emails. The dataset consists of 45% suspicious emails and 55% non-suspicious emails. Some examples from the dataset are given in Appendix.

The paper is organized as follows: Section II describes related work, whereas Section III explains various classification algorithms used for experimentation. Section IV discusses problem statement while Section V elaborates the proposed methodology. Section VI illustrates our experimental results and Section VII concludes the paper with future work.

## II. RELATED WORK

The research in the area of email analysis usually focuses on two areas namely: email traffic analysis and email content analysis. A lot of research has been conducted for Email traffic analysis [10],[11]. An email traffic analysis system manipulates the traffic part of the email to investigate the unusual behavior

[11] of suspicious individual. The traffic part of an email includes To, Carbon copy (Cc), Blind Carbon copy (BCc) and the Date fields. Email content analysis [11], [1], [22] on the other hand is the study of the unstructured part of the email such as the subject and body. Keila and Skillicorn [11] have investigated on the Enron [13]data set which contains email communications among employees of an organization who were involved in the collapse of the organization. The authors [1] have applied ID3 algorithm to detect suspicious emails by using keyword base approach and by applying rules. They have not used any information regarding the context of the identified keywords in the emails. S. Appavu & R. Rajaram [2] have applied association rule mining to detect suspicious emails with the additional benefits of classifying the (suspicious in terms of terror plots) emails further into specialized classes such as suspicious alert or suspicious info. This system decides whether the email can be classified as suspicious alert in the presence of suspicious keyword in the future tense otherwise only it is classified as suspicious info. The authors [14],[15] incorporated feature selection strategies along with classification systems. According to [15], by using feature selection methods one can improve the accuracy, applicability, and understandability of the learning process. Selvakuberan et al. [14]have applied filtered feature selection methods [16] on web page classification; according to their results the evaluator CfsSubsetEval yields better performance with search methods BestFirst, Ranker search, and Forward selection. Pineda-Bautista et al. [17] proposed a method for selecting the subset of features for each class in multi-class classification task. The classifiers that were used by the authors were Naive Baye's (NB) [6], k-Nearest Neighbors (k-NN) [17], C4.5 [19], and MultiLayer Perceptron (MLP). The authors trained the classifier for each class separately by using only the features of that particular class. Durant and Smith [20]have emphasized the use of a feature selection method for achieving accuracy of sentiment classification. They proposed to apply CfsSubsetEval with the BestFirst search method.

The ID3 (a type of decision tree) [21], algorithm is mostly used for email classification and content analysis systems [1],[2]. In the classification experiments for email spam filtering [21], decision tree classifiers outperform the other classifiers like SVM [23], neural networks [24], and others. SVM have also been applied for content extraction in the terrorism domain [25]. The method focuses on the context along with keywords. In general terms, features are collection of patterns on which the classification task is performed. Indeed, selecting an optimal set of features is generally difficult, both theoretically and empirically [25]. For the classification tasks, the proper strategy for feature selection is of utmost importance as emphasized in [27]. In this paper, we compare the evaluation of state of art decision tree, logistic regression, SVM, and Naïve Bayes [6] algorithms first without explicitly selecting features and then with various feature selection approaches. Our experiments show that choosing the most appropriate feature selection method can significantly improve the performance accuracy of existing state of the art classification algorithms for detection of suspicious emails.

## III. CLASSIFICATION ALGORITHMS

### A. Decision Tree

A decision tree [5],[19] consists of two types of nodes, namely; internal and external. Internal nodes correspond to attributes selected by decision tree algorithm for making decision at specific level of hierarchy. The branches coming out from these internal nodes are the values of that attribute. The attribute at top level of hierarchy in the tree has more power of classifying the instances of different classes. The external nodes in the tree correspond to the decision classes. Decision tree classifiers have some advantages over other classifiers, i.e., it is simple to build, its generated rules are easily interpretable by human and it is an inductive algorithm. Its accuracy can be very high, if an adequate training set is provided. There have been many decision tree algorithms developed through the time. Iterative Dochotomiser3 (ID3) [5] has remained the choice of data mining research community for many years. Beside its salient it has also some restrictions, i.e., ID3 can only deal with categorical attributes, it cannot handle missing values and it is not incremental. This led to the development of C4.5 [19] algorithm which can address the restrictions of ID3.

### B. Naive Bayes (NB)

Naïve Bayes [6] is a generative classification method that is based on Bayes theorem. It calculates the prior probabilities of each class and probabilities of each attribute in each class. It assumes that the probabilities of each attribute are independent of each other. At the time of classification it uses the prior probabilities of each class and the probabilities of the observed attributes. The class with highest probability is assigned to the instance being classified.

### C. Support Vector Machine (SVM)

SVM is a discriminative supervised machine learning technique of classification. SVM applies Vapnik's statistical learning theory [7] to train classifiers. SVM has some salient features for which it has been considered as state of art in the classification tasks. SVM has been used for text classification, hand written digit detection and many other classification tasks. Some of its unique features are: it can work well in a very high dimensional feature space, it uses only a subset of original training set to make decision boundary called support vectors and it is also suitable for non-linearly separable data (it uses kernel trick). Author [28] has described a number of features that explain why SVM is ideal for text classification tasks.

### D. Logistic Regression

Logistic Regression [7] belongs to the generalized linear model category of statistical models. It can predict a discrete outcome from a set of variables that may be categorical, numerical, continuous or dichotomous.

## IV. PROBLEM STATEMENT

The problem under consideration is to identify emails that contain suspicious contents indicating future terrorism events. We consider the task of suspicious email detection as a classification task. We start with a training set $T = \{e_1, e_2, e_3 \ldots e_m\}$ and class labels isSuspicious = {Yes, No}. Each email is given a label. The purpose is to formulate a model that learns from the training set and is able to classify a new email sample as either suspicious or non-suspicious.

We cannot deny the importance of email that is a major source of communication among most individuals and organizations, including terrorists and terrorist organizations. From this major source of communication, we can potentially locate evidence of future terrorist events. We propose a methodology to find clues about such events through email communication before those events take place. The proposed system first extracts useful features from the email body. If such features present in a certain combination, the email is marked as suspicious and the evidence of a potential future terrorist event is captured. If some features i.e., keywords are present but not others i.e., suspicious indicators, it may just be an email discussing past events, maybe condemning the events and so on. In the paper, we extracted the features (that are suspicious) along with the context. For example: an email body contains the message text as: "All the true Muslims condemn the terrorist attacks of 9/11." In the sentence, the keywords "terrorist" and "attack" are used but they do not indicate a future attack due to the presence of another feature "condemn." In the following section we present the methodology for problem under consideration.

## V. PROPOSED METHODOLOGY

Our proposed methodology uses machine learning techniques to detect the suspicious emails. It evaluates the performance of four classifiers with feature selection strategies. The algorithm for proposed methodology is given in the Fig. 1.

```
SED-FS(N)
[Initialize Classifiers] Classifier[4] ={Logistic, NB, ID3, SVM}
[Initialize Feature Selection] FS[10] = {WFS, CFS-BFS, CFS-GSS, CFS-RS, CSE-BFS, CSE-GSS, CSE-RS, IG-R, GR-R, Chi-R }
For( i = 1 to 4)
        For (j = 1 to 10)
                Apply FS[j]
                Classify using classifier[i]
                For(k = 1 to N)
                        If(y_{i,j} = y_k)
                                N_c=N_c+1
                End for
                Accuracy[i][j] = N_c/N*100
                Output = Accuracy[i][j]
        End for
End for
```

Fig. 1. Suspicious Email Detection Algorithm

The algorithm described in Fig. 1, illustrates the way suspicious email detection model works. In the algorithm, the variable FS is an array of 10 feature selection strategies and has the values as initialized for various feature selection schemes. WFS stands for Without Feature Selection, CFS-BFS stands for CfsSubSetEval and Best First Search method, CFS-GSS stands CfsSubSetEval and Greedy Stepwise Search method, CFS-RS stands for CfsSubSetEval and Rank Search method, CSE-BFS stands for ConsistencySubsetEval and BestFirst Search method, CSE-GSS stands for ConsistencySubsetEval and GreedyStepwise Search method while CSE-RS stands for ConsistencySubsetEval and Rank Search method, IG-R stands for InfoGain and Ranker search method, GR-R stands for GainRatio evaluation and Ranker search method where Chi-R stands for ChiSquare evaluation and Ranker search methods.

As the significance of feature selection strategy in the task of email classification has been identified, the next subsection discusses it analytically. In the following section, the system architecture is described to clarify how the feature selection presented is incorporated with the classifiers.

### A. Feature Selection

Feature selection is a way to select a subset of the original feature space. The number of features in the space affects the computation time and also the accuracy of the classifier. The key idea behind feature selection is to search a feasible subset of features by evaluating them, through some evaluators [14]. In this paper we focus on proper feature selection by which we could achieve relatively better performance of the required task even with the existing algorithms. Feature F is defined as a vector of K and I and it is the original feature space:

$$F = \{K_1, K_2, \ldots K_n, I_1, I_2, \ldots I_n\} \quad (1)$$

K is a vector of n keywords and I is a vector of indicators. Among the indicators some indicators make the email suspicious and some make the email non suspicious:

$$I = I_s + I_n \quad (2)$$

isSuspicious is a function over *K* and *I*:

$$\begin{aligned}
&\text{isSuspicious } (K,I) = \text{"Yes", if } (K=1 \text{ and } I_s = 1) \\
&\text{isSuspicious } (K,I) = \text{"Yes", if } (K=1 \text{ and } I_s = 1 \text{ and } I_n = 1) \\
&\text{isSuspicious } (K,I) = \text{"No", if } (K=1 \text{ and } I_n = 1) \\
&\text{isSuspicious } (K,I) = \text{"No", if } (K=1 \text{ and } I_s = 0 \text{ and } I_n = 1)
\end{aligned} \quad (3)$$

In the proposed approach we have not only used the terrorism domain keywords as features but also certain indicators such as the word 'condemn' as presented in the previous example. If a keyword is used in combination with a non-suspicious indicator, then it is not an indication of an upcoming terrorist event.

For our task, we have applied a supervised filtered feature selection method [16] because our task is a typical supervised machine learning classification task.

In the feature selection process we have applied

CfsSubsetEval, ConsistenceySubsetEval, InfoGain, GainRatio, and ChiSquare evaluators and BestFirst, GreedyStepwise and Ranker search methods. We have applied

*B. System Architecture*

The strategy used in the proposed system, for feature selection and generation is the one that it is kept outside the main classification engine, which contains different classification algorithms. This means that the feature selection is applied before classification. We used the well known open source data mining tool WEKA (Waikato Environment for Knowledge Analysis) [29]software for our feature selection and classification purposes.

Initially a text message is given as input and then feature selection strategies are applied. Extracted features are recorded in the ARFF file format which is the WEKA specific form of a Comma Separated Values (CSV) file. WEKA expects the input file to be in ARFF format. The communication between WEKA and our proposed system takes place with the help of data file exchanges. The original feature space is generated separately from WEKA with the help of a rule execution engine. The rest of the feature selection methods are applied manually from WEKA's filter feature selection. The email content is inspected and the rules are derived from the keywords and indicators in corresponding repositories. The rule execution engine executes these rules and the results of rule execution are used by the feature function factory to formulate the corresponding feature functions. The derived feature functions are kept in a feature function repository for future reuse. These features are applied on the decision tree, Naïve Bayes, SVM, and Logistic Regression classifiers to classify emails as suspicious or non suspicious. The described system architecture is shown in Fig. 2.

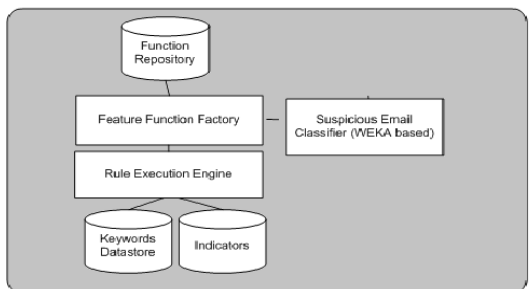

Fig. 2. System architecture

## VI. EXPERIMENTAL RESULTS

Our experiments were conducted using the steps defined in the algorithm given in Fig. 1. For conducting the experiments as mentioned before we used WEKA data mining tool. We have developed the terror email dataset from various sources like news groups. We also have used some emails that are used by authors [1],[2]. Some other emails are also dummy emails. The reason for developing such a dataset is because to the best of our knowledge, there is no such benchmark email dataset available in the counterterrorism domain.

For evaluating our experimental results we have used 10 fold cross validation. In this method the dataset is divided into 10 various combinations of evaluators and search methods and applied them on each of the four state of the art classifiers.

subsets and the algorithm runs in ten passes. In each pass one subset is used for testing and the rest nine of them are used as training sets. In each pass a new test set is selected and finally the average accuracy is returned. The accuracy A is measured as

$$A = \frac{1}{p} \sum_{i=1}^{p} ac_i \qquad (4)$$

where ac is the accuracy of correctly classified emails in pass $i$ and $p$ is the total number of passes. The experimental results show that the accuracy of the suspicious email detection task not only depends on the classifier itself but also on the feature selection strategy. Firstly, we conducted the experiments without using any feature selection strategy. This resulted in a relatively poor accuracy of the four algorithms. The results can be observed in Table II. Secondly, the experiments were conducted by applying feature selection strategies on each of the classification algorithms. Finally, the accuracies of the algorithms were compared with and without various feature selection methods. The results which are illustrated in TableII showed the highest accuracy. The results highlighted using bold face illustrates the second highest accuracy.

The accuracy of the logistic regression algorithm has been increased from 69.64% to 83.92%. Performance of the decision tree algorithm has been increased from 78.57% to 83.92%. The Naïve Bayes algorithm increased its performance from 69.64% to 78.57%. Finally, the SVM algorithm increased its performance from 73.21% to 80.35%.

The effect of each feature selection method on each of the algorithms, i.e. ID3, Naive Bayes, logistic regression and SVM is depicted in the graph Figures 3, 4, 5 and 6 respectively. It can also be observed from the results that an appropriate feature selection strategy greatly affects the performance of the logistic regression algorithm. ID3 is the best among them when no feature selection method is applied and also achieves maximum performance with feature selection. From the experiments it can be observed that with the feature selection method ConsistencySubsetEval (evaluator) and GreedyStepwise (search method) three of the four classifiers achieved the maximum performance – except the Naïve Bayes. Using feature selection methods like ChiSquare, InfoGain, and GainRatio with the Ranker search method resulted in the same accuracy of all the classifiers as without any feature selection method applied.

For the sake of space in graphs in Figures 3, 4, 5 and 6 we assign label to each feature selection scheme in Table I.

TABLE I: Feature selection schemes and corresponding labels

| Label | Feature Selection Scheme |
|---|---|
| 1 | Without Feature Selection |
| 2 | CfsSubsetEval, BestFirst Search |
| 3 | CfsSubsetEval, GreedyStepwise Search |
| 4 | CfsSubsetEval, RankSearch |

TABLE II: Experimental Results

| Method | Logistic regression | ID3 | Naïve Bayes | SVM linear |
|---|---|---|---|---|
| Without Feature Selection | 69.64% | 78.57% | 69.64% | 73.21% |
| CfsSubsetEval, BestFirst Search | 83.92% | 80.35% | 78.57% | 80.35% |
| CfsSubsetEval, GreedyStepwise Search | **83.92%** | **83.92%** | **76.78%** | **78.57%** |
| CfsSubsetEval, RankSearch | 75.00% | 75.00% | 76.78% | 73.21% |
| ConsistencySubsetEval, BestFirst Search | 82.14% | 80.35% | 78.57% | 80.35% |
| ConsistencySubsetEval, GreedyStepwise Search | <u>83.92%</u> | <u>83.92%</u> | <u>76.78%</u> | <u>80.35%</u> |
| ConsistencySubsetEval, Rank Search | 76.78% | 75.00% | 75.00% | 73.21% |
| GainRatio, Ranker Search | 69.64% | 78.57% | 69.64% | 73.21% |
| InfoGain, Ranker Search | 69.64% | 78.57% | 69.64% | 73.21% |
| ChiSquare, Ranker Search | 69.64% | 78.57% | 69.64% | 73.21% |
| 5 | ConsistencySubsetEval,BestFirst Search | | 8 | GainRatio, Ranker Search |
| 6 | ConsistencySubsetEval, GreedyStepwise Search | | 9 | InfoGain, Ranker Search |
| 7 | ConsistencySubsetEval, Rank Search | | 10 | ChiSquare, Ranker Search |

## VII. CONCLUSION AND FUTURE WORK

In this paper, we have presented suspicious email detection strategies using various classifiers and different feature selection methods. We concluded that in the specific task, the decision tree algorithm (ID3) outperformed the rest of the state of the art classifiers as Naïve Bayes, SVM, and logistic regression. After applying the appropriate feature selection strategy, the logistic regression algorithm also gave the maximum performance together with the decision tree algorithm. We also concluded that a feature selection strategy using ConsistencySubsetEval (as evaluator) and GreedyStepwise (as search method) achieves the maximum performance gain in terms of accuracy. In the future, we plan to also apply classifier based feature selection method for the specific task. We also plan to apply feature selection method on boosting algorithm for suspicious email detection task. At the moment, the experiments are conducted in relatively small dataset but in the future, we are planning to construct a larger dataset from terrorist statements derived from news groups, blogs, forums and terrorist websites.

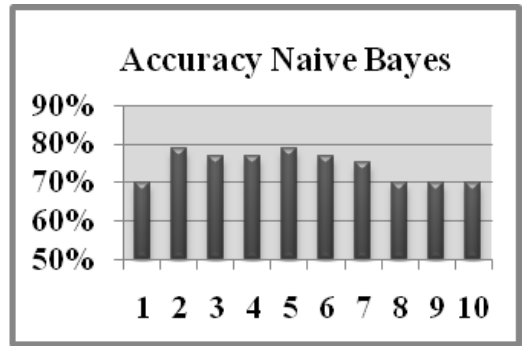

Fig. 4. Impact of feature selection in Naive Bayes

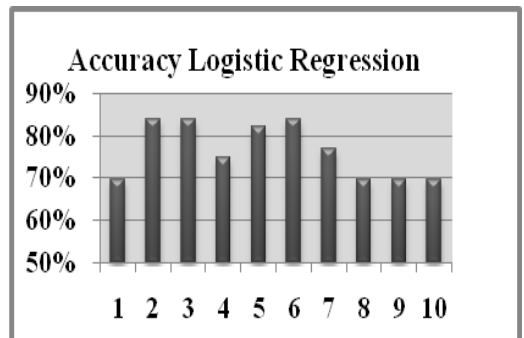

Fig. 5. Impact of feature selection in logistic regression

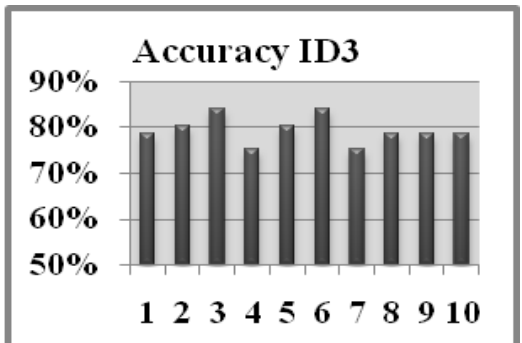

Fig. 3. Impact of feature selection in ID3

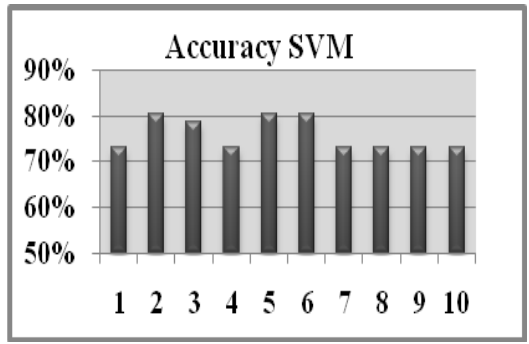

Fig. 6. Impact of feature selection in SVM

## APPENDIX

Some emails from the dataset

*Email 1.*

Dear Ali,

The incidence of 9/11 is really a big disaster. Many innocents become the victims of it. We all muslims do condemn it.

God bless everyone.

Regards,
Abid

*Email 2.*

Dear,

Hope you will be fine. What about preparations of our plan, I have completed all the preparations. This time there must be a big disaster. We will meet there exactly at 2 o'clock.

Regards
Anne

*Email 3.*

There is going to be blast in parliament building today. This time we want to kill big fishes Hahaha...

*Email 4.*

Government of India

This attack is a reaction to those actions which Hindus have taken since 1947 onwards. Now, there shall be no actions. There shall only be reactions, again and again.
These shall continue until we have avenged each and every atrocity.

**Sarwat Nizamani** received her B.Sc. (Hons.) and M.Sc. (Hons.) in Computer Science in the years 1998 and 1999 respectively from University of Sindh, Pakistan. The author received her Master's of Science in Robotic Engineering, from University of Southern, Denmark in 2011.

She worked as Research Associate from 2000-2003, then as lecturer from 2003-2007 and as Assistant Professor from 2007 to-date in University of Sindh, Pakistan. Since April 2010, she is PhD student, at University of Southern, Denmark. She has total six publications, out of which two are published in Springer Lecture Notes series and four in Conferences. Her field of research interest is machine learning, natural language processing, data mining, data structures and algorithm analysis.

**Nasrullah Memon** holds a PhD in Intelligence and Security Informatics (Computer Science and Engineering) from the Aalborg University Denmark. He has received Master in Software Development from University of Huddersfield, U.K. His research interests include machine learning, natural language processing, information retrieval, information extraction, open source intelligence and investigative data mining. He has published more than 90 research articles in journals/ conferences/ book chapters of national and international repute. He served as Editor-in-Chief of Social Network Analysis and Mining and affiliated with a number of international journals and conferences.

**Uffe Kock Wiil** is a Professor of Software Engineering and Technology at the Maersk Mc- Kinney Moller Institute, University of Southern Denmark. He holds a M.Sc. degree in Computer Engineering (1990) and a Ph.D. degree in Computer Science (1993) both from Aalborg University, Denmark. His research interest includes knowledge management, hypertext, computer supported cooperative work, software technology, and distributed systems. These research interests are currently being applied in three overall areas: counterterrorism, healthcare, and planning. He has published more than 150 research papers. His research papers have been cited more than 1300 times. He is currently serving on the editorial boards of the Elsevier Journal on Network and Computer Applications, the Springer Journal on Social Network Analysis and Mining, and the Hindawi ISRN Software Engineering Journal, and on the advisory board of the Springer Journal on Security Informatics.

**Panagiotis Karampelas** holds a PhD in Electronic Engineering from the University of Kent at Canterbury, UK. He also holds a Master of Science from the Department of Informatics, Kapodistrian University of Athens and a Bachelor degree in Mathematics from the same University. He has worked for 3½ years as an associate researcher in the Foundation for Research & Technology-Hellas (FORTH), Institute of Computer Science, and several years as a user interface designer and usability expert in several IT companies designing and implementing large-scale information systems. He has also participated in many European research projects and published a number of articles in his major areas of interests. He serves as an associate editor in the Social Network Analysis and Mining journal and reviewer in various scientific journals and conferences in his fields of interests. He teaches human computer interaction, programming and databases. Panagiotis Karampelas is also the Director of the Business and Information Technology Division of the Hellenic American University.